\documentclass{article}
\usepackage{ijcai25}
\usepackage{times}
\usepackage{soul}
\usepackage{url}
\usepackage[hidelinks]{hyperref}
\usepackage[utf8]{inputenc}
\usepackage[small]{caption}
\usepackage{graphicx}
\usepackage{amsmath}
\usepackage{amsthm}
\usepackage{booktabs}
\usepackage{algorithm}
\usepackage{algorithmic}
\usepackage[switch]{lineno}
\usepackage{pifont}
\usepackage{enumitem}
\usepackage{amsmath}
\usepackage{amssymb}
\usepackage{color}
\usepackage{xcolor}
\usepackage{booktabs}
\usepackage{multirow}
\usepackage{makecell}
\usepackage{colortbl}
\usepackage{adjustbox}
\usepackage{tikz}
\usepackage{subfigure}
\usepackage{enumitem}
\usepackage{bbding}
\usepackage{array}

\newcommand{\modelname}{MimbFD}

\title{Mitigating Message Imbalance in Fraud Detection with Dual-View \\ Graph Representation Learning}

\author{
  Yudan Song$^{1,2}$\textsuperscript{†} \and
  Yuecen Wei$^{3,4}$\textsuperscript{†} \and
  Yuhang Lu$^{1,2}$\and
  Qingyun Sun$^{3}$\and
  Minglai Shao$^{5}$\and \\
  Li-e Wang$^{1,2}$\thanks{Co-corresponding Authors.} \and
  Chunming Hu$^{3,4}$ \and
  Xianxian Li$^{1,2}$\and
  Xingcheng Fu$^{1,2}$\footnotemark[1]
  \affiliations
  $^1$Key Lab of Education Blockchain and Intelligent Technology, Ministry of Education, \\Guangxi Normal University, Guilin, China\\
  $^2$Guangxi Key Lab of Multi-Source Information Mining and Security, \\Guangxi Normal University, Guilin, China\\
  $^3$School of Software, Beihang University, Beijing, China \\
  $^4$SKLCCSE, School of Computer Science and Engineering, Beihang University, China \\
  $^5$School of New Media and Communication, Tianjin University, China
  \emails
        \{songyudan, lyh0620\}@stu.gxnu.edu.cn, 
        \{wanglie,lixx,fuxc\}@gxnu.edu.cn, 
        \{weiyc,sunqy,hucm\}@buaa.edu.cn, 
        shaoml@tju.edu.cn\\
}

\begin{document}

\maketitle

% \footnotetext{† Co-first Authors, * Co-corresponding Authors.}
% \footnotetext{* Co-corresponding Authors.}

\begin{abstract}
Graph representation learning has become a mainstream method for fraud detection due to its strong expressive power, which focuses on enhancing node representations through improved neighborhood knowledge capture. 
However, the focus on local interactions leads to imbalanced transmission of global topological information and increased risk of node-specific information being overwhelmed during aggregation due to the imbalance between fraud and benign nodes. 
In this paper, we first summarize the impact of topology and class imbalance on downstream tasks in GNN-based fraud detection, as the problem of imbalanced supervisory messages is caused by fraudsters' topological behavior obfuscation and identity feature concealment.
Based on statistical validation, we propose a novel dual-view graph representation learning method to mitigate \textbf{M}essage \textbf{imb}alance in \textbf{F}raud \textbf{D}etection (\modelname).
Specifically, we design a topological message reachability module for high-quality node representation learning to penetrate fraudsters' camouflage and alleviate insufficient propagation. 
Then, we introduce a local confounding debiasing module to adjust node representations, enhancing the stable association between node representations and labels to balance the influence of different classes. 
Finally, we conducted experiments on three public fraud datasets, and the results demonstrate that {\modelname} exhibits outstanding performance in fraud detection.

\end{abstract}

\section{Introduction}
With the rapid development of intelligent technology, various applications facilitate people's lives, such as financial applications~\cite{Xu21AAAI}, social media~\cite{PengZLCPY23,hao2024multi}, and review systems~\cite{DhawanGK019}. However, factors such as limited access to advanced knowledge by vulnerable groups (benign users) cause malicious fraudsters to exploit information gaps to execute successful fraudulent tactics~\cite{Song24}. This widespread phenomenon severely affects the order on various platforms, bringing significant attention to fraud detection technologies~\cite{Survey}. Thus, fraud detection serves as an effective means for early warning, timely intervention, and informed decision-making~\cite{Impact}.

\begin{figure}[t]
\centering
\includegraphics[width=0.48\textwidth]{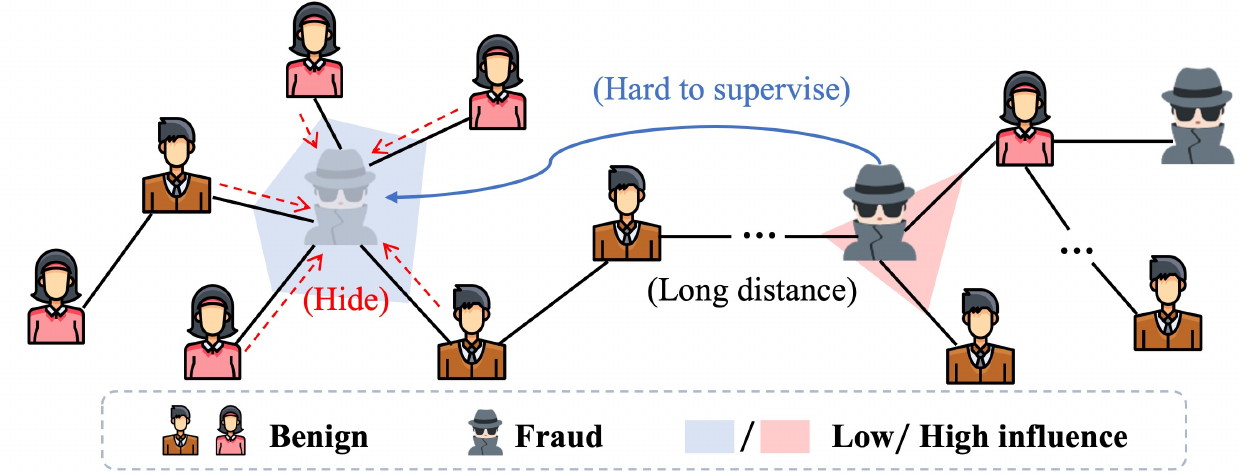} 
\caption{
Fraudster distribution patterns. 
The concealment strategies of fraudsters in different topological spaces result in varying levels of influence and lead to unreachable identity supervisory messages.
}
\label{fig:motiv}
\end{figure}

In recent years, thanks to the strong expressive power of Graph Neural Networks (GNNs)~\cite{Wei25aaai,Wei24aaai,Fu25AAAI}, an increasing number of works~\cite{Survey} have adopted GNNs for fraud detection research. 
Some works focus on the fraudster's camouflage behavior, where fraudsters mimic the behavior patterns of benign users and interact with them, deceiving the model into classifying them as benign~\cite{GAGA,COFD}. 
To uncover such camouflage, these methods emphasize differentiating fraudsters from benign users regarding node representation, thereby enhancing the model's ability to identify anomalies~\cite{CARE-GNN,FRAUDRE,PC-GNN,ConsisGAD}. 
%Additionally, the others tackle the problem from the perspective of graph structure types or properties, modeling graph data as heterogeneous graphs to learn representations and leveraging class imbalance properties to draw the model's attention to less frequent anomalous classes, thereby improving detection performance~\cite{PC-GNN,ConsisGAD}.
However, they often focus only on local interaction features, overlooking critical long-distance information and neglecting to consider the impact of the fusion of diverse supervised information when dealing with locally confounding messages.
In such cases, some fraudulent activities may be overlooked due to the lack of a global perspective. 
Therefore, starting from the topological distances in the graph, capturing relationships and patterns between long-distance nodes is promising for a better understanding of fraudulent behavior.
Next, we discuss the impact of class and topological imbalance on supervisory message transmission in fraud scenarios, as shown in Fig.~\ref{fig:motiv}. 
\textit{Firstly}, fraudsters are dispersed across different communities and exhibit an imbalance in the transmission of topological information to influence their neighbors due to the diversity of the topological structure, either to expand the impact of fraud or to achieve camouflage.
\textit{Secondly}, at the data level, fraudsters are the minority, while benign users are the majority. This class imbalance leads to the supervisory message of the minority class being overwhelmed during the message-passing process in GNN, causing severe information aggregation drift.

\textbf{Challenges.} 
Although GNN-based fraud detection has achieved nice performance, it still faces the following two challenges in solving the problem of inadequate representation of the imbalanced supervisory message:
$(1)$ Graph representation learning (GRL) often distorts when aggregating supervisory messages from distant topological locations. Furthermore, due to the topological message imbalance brought by fraudsters' preferences, nodes fail to learn key useful representations. Therefore, \textit{the challenge is to fully acquire the messages from the global topology to mitigate the conflict between the intrinsic features of nodes and the representations of local neighborhoods.}
$(2)$ Existing methods tend to adjust the local neighborhood to avoid unnecessary noise. Due to the extreme class imbalance between fraudsters and benign users, such methods magnify information bias and provide favorable conditions for fraudsters to camouflage. \textit{The challenge lies in how to establish an essential correlation between node representation and labels and enhance their connection among massive messages.}

\textbf{Contributions.} 
We first perform statistical analysis on fraudsters' topological behavior obfuscation and identity feature concealment intentions, uncovering the supervisory message imbalance phenomenon in GNN-based fraud detection.
To address these challenges, we propose a novel dual-view graph representation learning method to mitigate \textbf{M}essage \textbf{imb}alance in \textbf{F}raud \textbf{D}etection, named {\modelname}. 
Specifically, we uncover fraudsters' camouflage by capturing information that is strongly correlated with the network topology and offer guidance for mitigating topological imbalance.
We then introduce a mechanism to de-correlate confounding associations in local messages, aiming to emphasize representations that reflect nodes' intrinsic information and their association with labels. 
Finally, empirical results on three datasets confirm the method's effectiveness in alleviating supervisory message imbalance.
In conclusion, our contributions are summarized as follows:

\begin{itemize}
    \item We are the first to categorize the imbalance problem in fraud detection from a dual view as an imbalance in supervisory messages in GRL, highlighting the importance of mitigating informational gaps. 
    \item We propose a dual-view graph representation method for fraud detection, 
    adjusting the propagation of supervisory messages while enhancing the connection between critical information that determines node identity and labels. 
    \item Comprehensive experiments demonstrate that {\modelname} significantly mitigates the effect of imbalance and improves fraud detection performance. 
\end{itemize}
\section{Related Work}

\subsection{Imbalanced Learning On Graphs}

The direct application of GNNs on imbalanced datasets results in outcomes that are biased toward the majority class, which is why most solutions~\cite{igl,Fu23} to unbalanced problems in graph learning focus on addressing class imbalance.
Common methods include generating nodes and reweighting techniques. 
Generative node methods primarily focus on generating high-quality minority class samples to enhance the learning of the minority class~\cite{GRAPHENS,GraphSMOTE,GraphSHA,ImGAGN}.
Reweighting~\cite{TAM}, on the other hand, adjusts the classification boundaries of individual classes by designing specialized loss functions. 
Another significant area in graph imbalance learning is topological imbalance. 
\cite{GraphPatcher} focuses on the local topological imbalance of nodes and achieves enhancement of nodes by repairing low-degree nodes.
ReNode~\cite{ReNode} is the first work to address the problem of global topological imbalance, addressing the issue of information conflict among nodes through adaptive reweighting. 
Similarly, ~\cite{PASTLE} advances the exploration of topological imbalance by optimizing information propagation paths via position-aware augmentation.
However, these works lack an in-depth analysis of fraudster behavior. As a result, they are limited against camouflage. This limitation makes them difficult to apply directly to fraud detection.

\subsection{Fraud Detection Based on GNNs}
The powerful expressive capability of GNNs has garnered significant research interest. In recent years, numerous graph-based fraud detection methods have been proposed, which can generally be categorized into two main types based on their approach to adjusting neighborhood structures: local adjustment and high-order capture.
\textbf{Local adjustment}: 
CARE-GNN~\cite{CARE-GNN} leverages reinforcement learning to guide the identification of similar nodes within a neighborhood.
FRAUDRE~\cite{FRAUDRE} emphasizes the differences between normal users and fraudsters in node pairs to strengthen the model's ability to detect fraudulent behavior.
However, these works mainly focus on the attributes of local neighborhoods and tend to ignore the critical information from distant nodes. 
\textbf{High-order capturing}:
PC-GNN~\cite{PC-GNN} introduces a node sampler and a label-aware neighbor selector, and additionally connects the fraudsters that are not directly linked to one another.
GAGA~\cite{GAGA} employs group aggregation for both first-order and second-order neighborhoods to enhance node representation.
COFD~\cite{COFD} adjusts node neighborhoods by comparing first-order neighborhoods with second-order neighborhoods.
While these methods consider the capture of high-order neighborhoods, they remain limited to small-scale mining of homophily to enhance node representations. The supervisory message available in the graph topology is not sufficiently leveraged in these works.

\begin{figure*}[ht] 
    \subfigure[Amazon.]{    \begin{minipage}{0.32\textwidth}
        \centering      
        \includegraphics[width=1\textwidth]{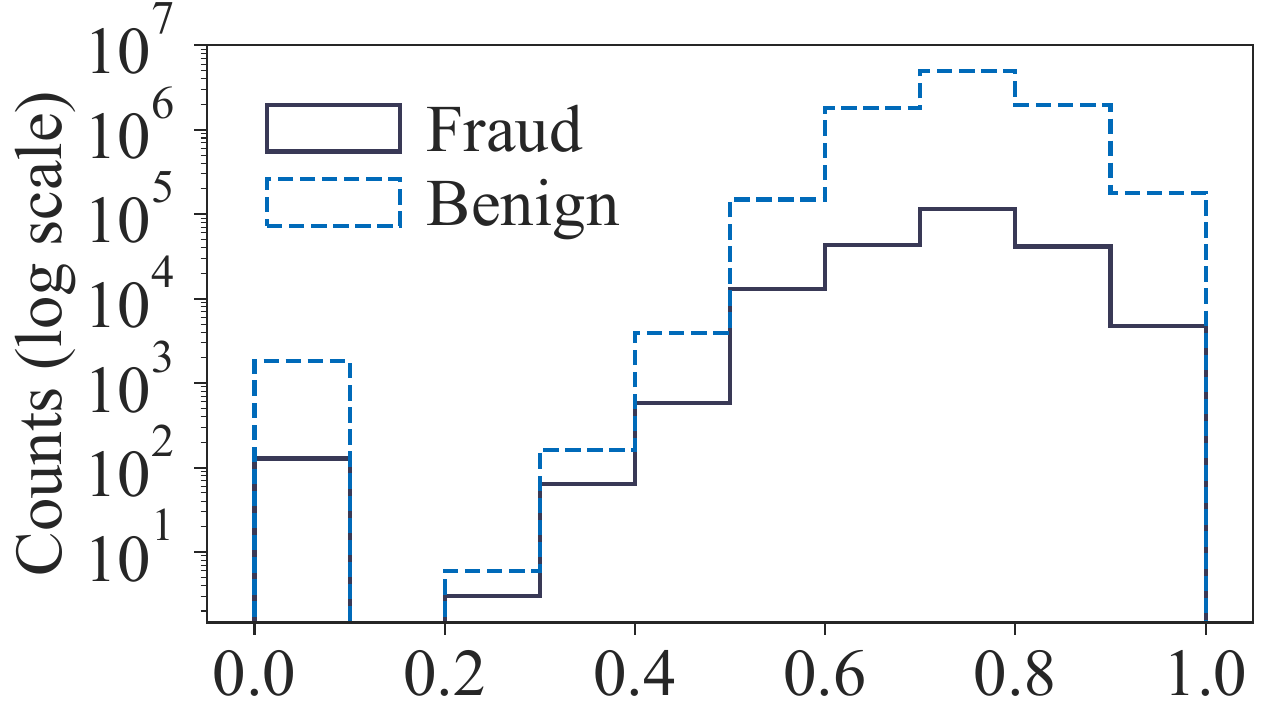}
    \end{minipage}}
    \subfigure[YelpChi.]{  
    \begin{minipage}{0.32\textwidth}
        \centering
        \includegraphics[width=1\textwidth]{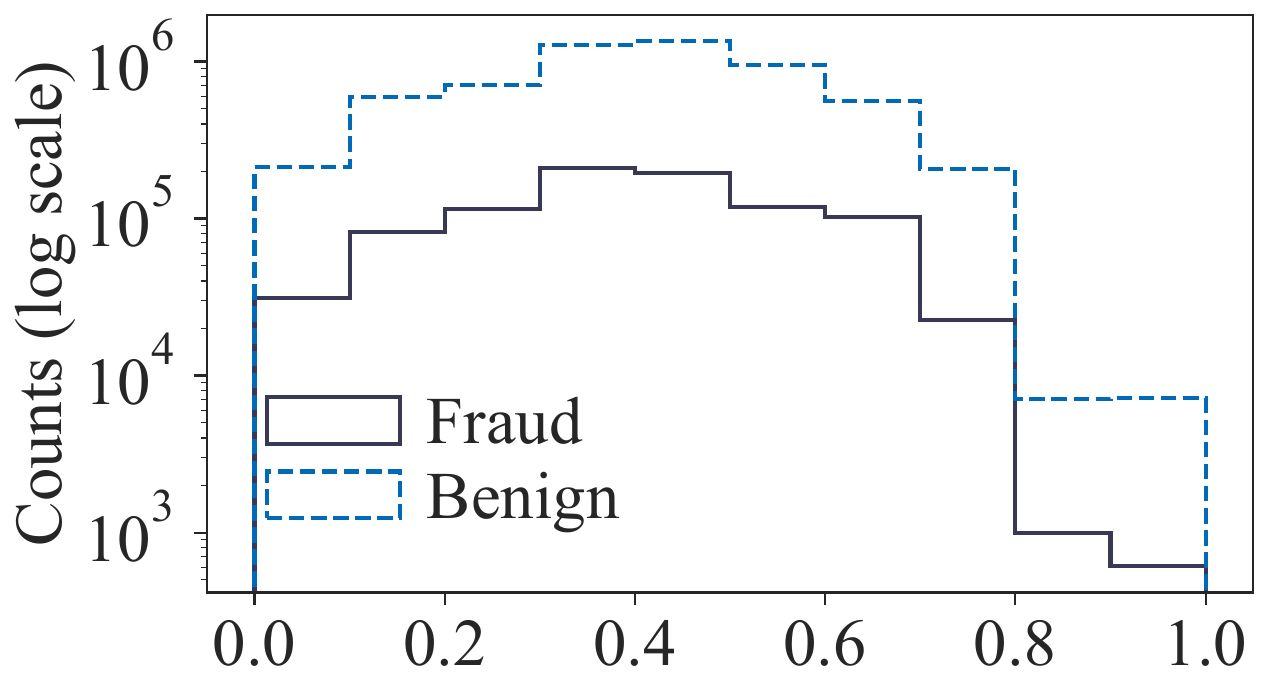}
    \end{minipage}
    }
    \subfigure[Comp.]{  
    \begin{minipage}{0.32\textwidth}
        \centering
        \includegraphics[width=1\textwidth]{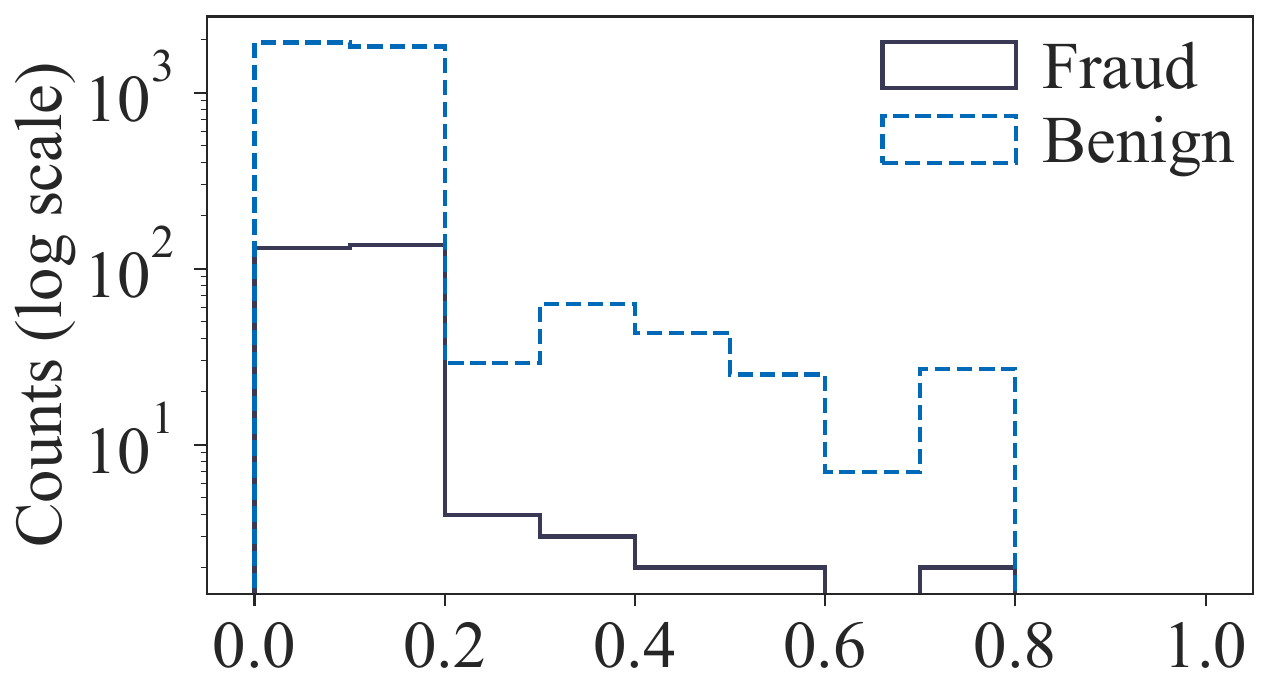}
    \end{minipage}
    }
\centering
\caption{Neighborhood distributions under datasets with different closeness centrality.}
\label{Evidence1}
\end{figure*}

\begin{figure*}[ht] 
    \subfigure[Amazon.]{    \begin{minipage}{0.32\textwidth}
        \centering      
        \includegraphics[width=1\textwidth]{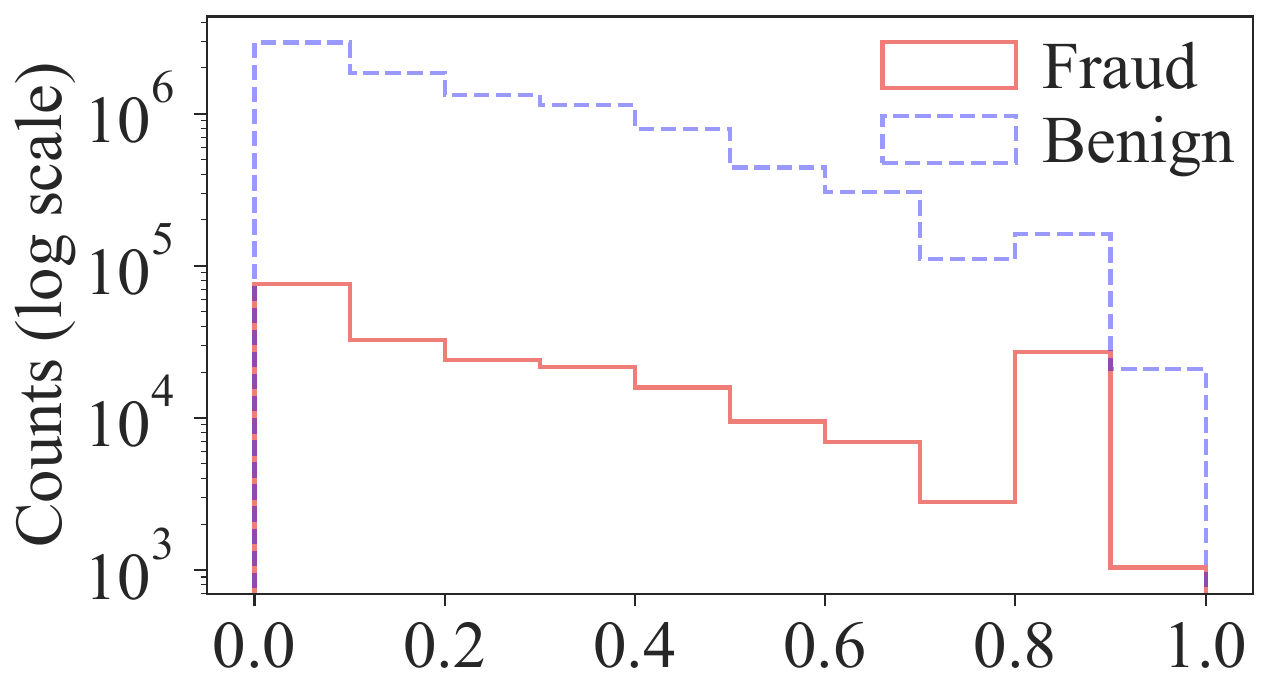}
    \end{minipage}}
    \subfigure[YelpChi.]{  
    \begin{minipage}{0.32\textwidth}
        \centering
        \includegraphics[width=1\textwidth]{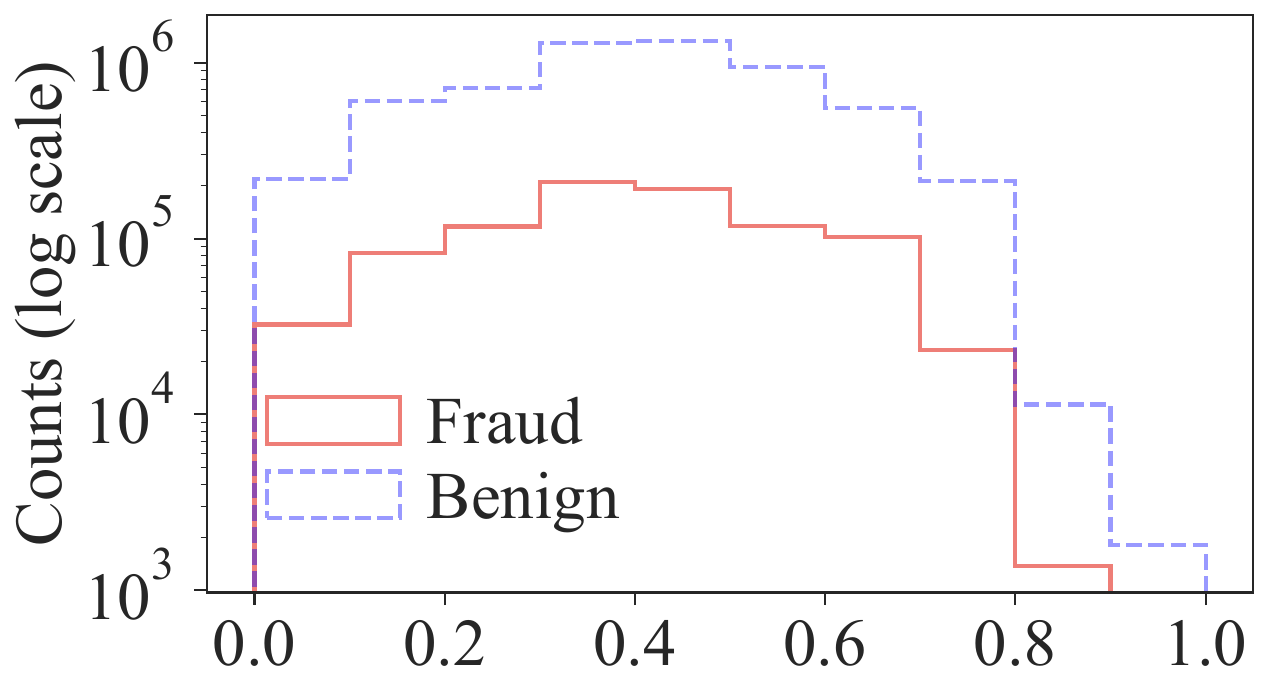}
    \end{minipage}
    }
    \subfigure[Comp.]{  
    \begin{minipage}{0.32\textwidth}
        \centering
        \includegraphics[width=1\textwidth]{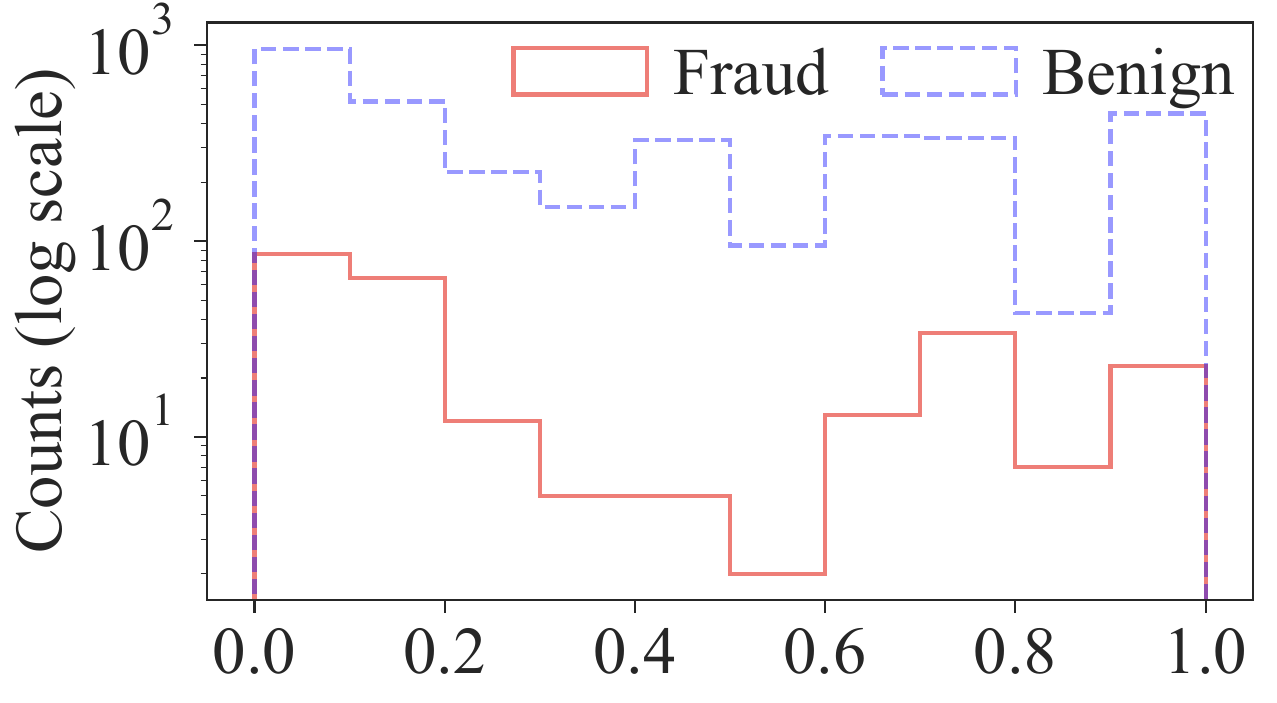}
    \end{minipage}
    }
\centering
\caption{Neighborhood distributions under datasets with different degree centrality.}
\label{Evidence2}
\end{figure*}

\section{Preliminary}

\subsection{Definition}
\textbf{Fraud detection on multi-relation graph}
Given a graph $G=\{\mathcal{V},\mathbf{X},\{{{\mathcal{E}}_{r}}\}|_{r=1}^{R},\mathcal{Y}\}$,which ${{v}_{i}}\in \mathcal{V}$ denotes the node, $\mathbf{X}$ denotes the features of the node, $\mathcal{R}$ denotes the kind of relationship, ${{e}_{i,j,r}}\in \mathcal{E}$ denotes the two nodes ${{v}_{i}}$ and ${{v}_{j}}$ are connected under the relationship ${r}$.
Graph-based fraud detection is a semi-supervised binary classification task. 
Only a small number of nodes are labeled (denoted by $\mathcal{Y}$) on the graph $G$. Each node has a binary label. ${{y}_{i}}=0$ denotes a benign user, ${{y}_{i}}=1$ denotes a fraudster.

\subsection{Problem Analysis}
To verify the existence of topological imbalance, we conducted data statistics on three public datasets. This helped us investigate the potential causes of the topological imbalance issue in fraud detection.

\noindent\textbf{Topological Behavior Obfuscation}
Closeness centrality (CC)~\cite{close} measures the average shortest path distance from a given node to all other nodes in the network, reflecting the speed at which the message spreads from that node to others. 
As illustrated in Fig.~\ref{Evidence1}, we analyze the distribution of fraudulent and benign neighbors for benign users with different CC values. 
A higher CC value indicates faster user information exchange in the social network.

On YelpChi and Comp, benign users with lower CC tend to have more fraudster neighbors. On Amazon, as CC increases, the number of both types of neighbors also increases, but the difference between their distributions becomes larger.
It suggests that benign users with lower CC are more likely to be surrounded by fraudsters, limiting the spread of fraud-related signals to a small area. In contrast, benign users with higher CC have more benign neighbors, and the number of fraudster neighbors grows more slowly, making it harder for fraud-related signals to spread.
Overall, fraudsters avoid connecting too much with core benign users to prevent clear fraud patterns in the network structure. As a result, their activities are often limited to areas where information spreads slowly. This makes it hard for fraud-related signals to reach wider regions, contributing to topological imbalance.

\noindent\textbf{Identity Feature Concealment}
Degree centrality (DC)~\cite{Degree} quantifies the importance of a node in the network based on the number of direct connections it has with other nodes, indicating its direct influence and connectivity. 
As shown in Fig.~\ref{Evidence2}, we analyze the distribution of fraudulent neighbors and benign neighbors of benign users with different DC values. 
A higher DC indicates more user connections within the network.

Fraudsters are mostly found near benign users with low DC, while benign users with high DC are surrounded by more benign neighbors.
Even when a fraudster connects to the benign users with high DC, they can still hide their identity effectively, as their information become diluted.
As a result, the spread of fraud-related information remains limited in scope, contributing to topological imbalance.

\section{Methodology}
\begin{figure*}[t]
\centering
\includegraphics[width=1\textwidth]{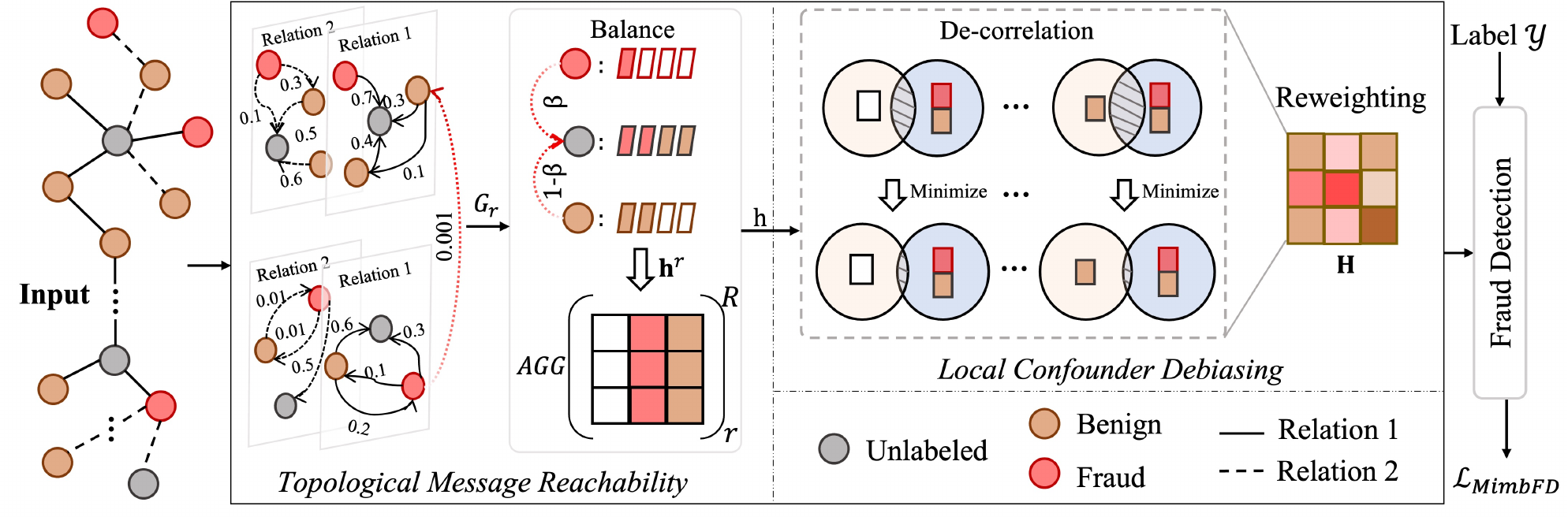} 
\caption{{\modelname} consists of two modules: (1) TMR: Utilizing the balancing factor to assist nodes in adapting to the effects of the propagation of labeled nodes. (2) LCD: Analyze the variables in node representations to amplify the key message associated with the labels.}
\label{fig:framework}
\end{figure*}

Our goal is to address the challenges in supervised message passing for fraud detection, which are impacted by both topological imbalance and class imbalance.  For topological imbalance, we introduce the topological message reachability (TMR) module, which enables global topology sensing to adjust the impact of long-distance messages received by nodes fully. To tackle class imbalance, we propose the local confounding debiasing module (LCD), which amplifies the weights of label-related variables while mitigating the discrimination bias introduced by the neighborhood.
The overall architecture of {\modelname} is illustrated in Fig.~\ref{fig:framework}.

\subsection{Topological Message Reachability}
To adaptively adjust the information received by nodes regarding the supervisory message of different classes of identity nodes at a distance. 
This adjustment enhances node connections within the same class and improves the ability to discriminate between benign users and fraudsters. 
To enable the unknown labeled nodes to adjust the supervisory information from two distinct classes of nodes, we first capture the diffuse impact of the supervisory message from the labeled nodes on the global topology. 
% To achieve this, 
Thus, we introduce Group PageRank~\cite{GPR} to measure the propagation influence of labeled nodes on the topology.
Specifically, under each relationship graph, the impact matrix corresponding to each class is calculated as follows:
\begin{equation}\label{eq:GPR(c)}
\begin{aligned}
g^{r}_{gpr}\left ( c \right )  = 
\left ( 1- \alpha^{r}  \right ) A^{'}_{r} g^{r}_{gpr}\left ( c \right ) +  \alpha^{r} I^{r}_{c},
\end{aligned}
\end{equation}
where $c$ represents the labeled categories $0$ and $1$ of the labeled nodes, $\alpha^{r}$ denotes the probability that class $c$ is restarted by a randomized wandering under a relational subgraph; $ A^{'}_{r}= A_{r}D^{- 1}$, $A$ denotes the adjacency matrix of the subgraph of relation $r$ and $D$ is the degree matrix. Additionally, $I^{r}_{c}\in  R^{n}$ is the transmission vector, defined as follows:
\begin{equation}\label{eq:I(c)}
\begin{aligned}
I^{r}_{c} =\begin{cases}\frac{1}{\left |\mathcal{V}_{L}^{c}   \right | },
  & \text{ if } y_{i} = c\\
  $0$,& otherwise
\end{cases}
\end{aligned}
\end{equation}
where $\mathcal{V}_{L}^{c}$ is the number of nodes that have labeled class $c$. 
%For $c=0$ and $c=1$, $I^{r}_{c}$ represents the degree of supervised information dissemination for benign users and fraudsters, respectively.
$I^{r}_{c}$denotes the supervised dissemination degree for benign ($c=0$) and fraudulent ($c=1$).

After calculating the corresponding information propagation matrix for each class, we connect them. The final impact propagation matrix is as follows:
\begin{equation}
\begin{aligned}
G^{r} = \alpha^{r}  \left ( E-\left ( 1-\alpha^{r} A^{'}_{r}  \right )   \right ) ^{-1}I^{*}_{r} ,
\end{aligned}
\end{equation}
where $E$ is the unit matrix and $I^{*}_{r}$ is the composition of the collocation of all classes $I^{r}_{c}$.
To some extent, the values in the matrix reflect the influence of the labeled nodes in the global topology. 
A higher value indicates a greater ability of the node to disseminate information throughout the topology.

To highlight the importance of nodes in the global topology message, we capture the neighborhood of the central node at a fine-grained level. For labeled neighbor nodes, weighted using normalized group PageRank values:
\begin{align}
\mathbf{p^{r}} &= softmax\left ( G^{r}  \right ),\\
\mathbf{h}_{be,j}^{\left ( l \right ), r} &= \sum _{v_{j}\in N_{be}\left ( v_{i}  \right )}\mathbf{p}_{be}^{r}\mathbf{h}_{j}^{\left ( l-1 \right ), r},\\
\mathbf{h}_{fr,j}^{\left ( l \right ), r} &= \sum _{v_{j}\in N_{fr}\left ( v_{i}  \right )}\mathbf{p}_{fr}^{r}\mathbf{h}_{j}^{\left ( l-1 \right ), r},
\end{align}
where $\mathbf{h}_{be,j}^{\left ( l \right ), r}$ and $\mathbf{h}_{fr,j}^{\left ( l \right ), r}$ are weighted representations of benign users and fraudsters, respectively.

For nodes in the neighborhood with unknown labels, we must address the issue of biasing them towards receiving information from any specific class of nodes. After all, information propagation in a topology inevitably involves some conflicts.
Therefore, we aim to distinguish the relationship between the unknown labeled nodes and the two types of known labeled nodes. 
Influence representations are assigned to unlabeled nodes as discriminative guidance, as follows:
\begin{equation}\label{eq:unbe(j)}
\begin{aligned}
\mathbf{h}_{un,be,j}^{\left ( l \right ), r} = \sum _{v_{j}\in N_{un}\left ( v_{i}  \right )}\mathbf{p}_{be}^{r}\mathbf{h}_{j}^{\left ( l-1 \right ), r},
\end{aligned}
\end{equation}
\begin{equation}\label{eq:unfr(j)}
\begin{aligned}
\mathbf{h}_{un,fr,j}^{\left ( l \right ), r} = \sum _{v_{j}\in N_{un}\left ( v_{i}  \right )}\mathbf{p}_{fr}^{r}\mathbf{h}_{j}^{\left ( l-1 \right ), r},
\end{aligned}
\end{equation}
where $\mathbf{h}_{un,be,j}^{\left ( l \right ), r}$ and $\mathbf{h}_{un,fr,j}^{\left ( l \right ), r}$ 
denote the influence representations of benign users and fraudsters on unlabeled nodes.

We then introduce a learnable adjustment factor to help the unknown labeled nodes adaptively combine these two representations in relation to themselves. The node's own representation of these two identities carries selective associations. To restore the association between the nodes, the adjustment factor is computed as follows:
\begin{equation}\label{eq:beta}
\begin{aligned}
\beta =\sigma \left ( \mathbf{h}_{i}^{\left ( l-1 \right ) }  \right ) ,
\end{aligned}
\end{equation}
where $\sigma \left ( \cdot \right )$ is the activation function.
The final neighborhood representation of the unknown labeled node is obtained by a weighted combination of the two representations using the adjustment factor, as follows:
\begin{equation}\label{eq:un}
\begin{aligned}
\mathbf{h}_{un,j}^{\left ( l \right ), r} = 
\beta \mathbf{h}_{un,fr,j}^{\left ( l \right ), r}+  \left ( 1-\beta  \right ) \mathbf{h}_{un,be,j}^{\left ( l \right ), r}.
\end{aligned}
\end{equation}

After obtaining three different types of node representations in the neighborhood, we integrate them with the central node's own representation. Simultaneously, the integration of the representations within each relationship subgraph is carried out. The final node representation is as follows:
\begin{equation}\label{eq:f}
\begin{aligned}
\mathbf{h}_{i}^{\left ( l \right )} = AGG_{all}^{l} \left [ MLP\left ( \mathbf{h}_{i}^{\left ( l-1 \right ),r} \right ) ,
\mathbf{h}_{be,j}^{\left ( l \right ), r},
\mathbf{h}_{fr,j}^{\left ( l \right ), r},
\mathbf{h}_{un,j}^{\left ( l \right ), r} \right ] _{r=1}^{R}.
\end{aligned}
\end{equation}

\subsection{Local Confounder Debiasing}
Due to class imbalance, the node representations obtained through message passing contain a significant amount of information about normal users.
Specifically, the node representation contains excessive confounding information, making it difficult for the model to identify key factors for determining identity. 
Thus, the model may favor the assumption that the node is a ``benign user." According to stable learning~\cite{stable}, node representations can be categorized into stable and unstable variables. 
Stable variables are those key factors that determine the labeling categories, regardless of the environment. 
Therefore, we aim to perform a de-correlation between variables in the node representation, identifying stabilizing variables that will strengthen the connection between the nodes and their corresponding labels.

For the node representation, we employ the de-correlation term~\cite{Debias} to ensure that the node variables are as independent from each other as possible. This helps alleviate spurious correlations between variables and labels. It is defined as follows:
\begin{equation}\label{eq:LCD}
\begin{aligned}
 \mathcal{L}_{LCD}= & \sum_{j=1}^{p}\left(\gamma  ^ { \mathrm { T } } \cdot  \mathbf{h}_{i,m}^{\mathrm{T}} \left |  
 \Lambda_{\mathbf{w}} \mathbf{h}_{i,-m} / n\right.\right. \\
& \left.\left.- \mathbf{w} / n \cdot \mathbf{h}_{i,-m}^{\mathrm{T}} \mathbf{w} / n\right |  
\right)^{2} \\
& +\frac{\lambda_{1}}{n} \sum_{i=1}^{n} \mathbf{w}_{i}^{2}+\lambda_{2}\left(\frac{1}{n} \sum_{i=1}^{n} \mathbf{w}_{i}-1\right)^{2}, \\
& \text { s.t. } \mathbf{w} \geq 0
\end{aligned}
\end{equation}
where $\mathbf{h}_{i,m}$ denotes the node representation $m$ of the variables in $\mathbf{h}_{i}$, and $\mathbf{h}_{i,-m}$ is the node representation $\mathbf{h}_{i}$ of all the variables in $\mathbf{h}_{i}$ except $m$. The first term represents the computation of the differences between different variables in the node representation and applies weights $\gamma$ to these variables to amplify their association with the true identity of the node.
The second term indicates stability by enforcing regularization.
The third serves to prevent the sample weights from going to zero.
The $\text { s.t. } \mathbf{w} \geq 0$ is a non-negative constraint.
To reduce computational complexity, this regularization is used at the last layer of the linear transformation.

Overall, the significance of de-correlation lies in analyzing the relationships between pairs of variables and reweighting each variable until the objective is minimized. This process results in better tuning of the node representation, strengthening its intrinsic connection with the labels.

\subsection{Training}
After iterative multiple layers, we take the output of the last layer as the final representation of the node $\mathbf{H}_{i} =\mathbf{h}_{i}^{L}$. We adopt the cross-entropy loss as the loss function for the classification result, which is defined as follows:
\begin{equation}\label{eq:GNNLOSS}
\begin{aligned}
{{\mathcal{L}}_{GNN}}=\sum\limits_{v\in V}{-\log ({{y}_{v}}\cdot \sigma (\textsc{MLP}({\mathbf{{H}
}_{i}})))}.
\end{aligned}
\end{equation}

Combined with the loss function in the local confounding debiasing module, we define the loss of {\modelname} as:
\begin{equation}\label{eq:totalLOSS}
\begin{aligned}
{\mathcal{L}}_{\modelname}={{\mathcal{L}}_{GNN}}+{\eta  }{{\mathcal{L}}_{LCD}},
\end{aligned}
\end{equation}
where ${\eta }$ is the balance parameter.

\begin{table*}[ht]
\centering
\resizebox{\textwidth}{!}{
\begin{tabular}{c|ccc|ccc|ccc}
\toprule

\multirow{2}{*}{\raisebox{-1ex}{Model}}
& \multicolumn{3}{c|}{Amazon} 
& \multicolumn{3}{c|}{YelpChi} 
& \multicolumn{3}{c}{Comp}

\\ \cmidrule{2-4} \cmidrule{5-7} \cmidrule{8-10}

&  AUC &  Rec &  F1 &  AUC &  Rec &  F1 &AUC &  Rec &  F1 
\\ \midrule

GCN   &85.83±0.06   &75.15±0.20   &55.26±0.96   &52.23±0.28   &54.20±1.37   &55.96±0.31 &55.03±0.11   &50.00±0.0   &47.24±0.0       \\
GAT    &81.02±0.49   &54.88±0.31   &64.64±0.38   &55.40±1.04   &48.00±0.0  &46.14±0.0  &55.03±0.0  &50.00±0.0  &47.24±0.0        \\
GraphSAGE    &73.67±0.17   &68.23±0.23   &56.96±0.82   &50.82±1.13   &50.08±0.30   &51.86±0.88  & 53.81±0.16 &45.69±0.56   &46.65±0.51      \\
\midrule
ReNode   &40.52±27.39   &25.68±11.2   &59.7±3.9   &-   &-   &- &51.18±3.62   &43.42±16.66   &43.2±6.3       \\
TAM   &50.19±0.88   &59.83±0.8   &61.60±0.75   &57.28±0.24   &53.75±0.24   &53.98±0.35 &53.08±0.23   &50.37±0.14   &48.92±0.47       \\
\midrule

CARE-GNN    &93.49±0.36  &89.13±0.10  &88.83±0.22  &76.85±0.07  &70.36±0.03  &60.99±0.10 &54.16±0.52   &46.84±0.4  &46.78±0.19      \\
GAGA       &\underline{95.27±0.24}  &87.95±0.70  &\underline{91.32±0.60}  &\textbf{92.99±0.42}  &\underline{80.95±1.18}  &\textbf{80.15±0.83}  &\underline{59.87±0.69}   &\underline{55.42±4.25}  &45.32±2.02    \\
COFD       &94.84±1.10  &\underline{89.52±1.30}  &88.64±1.42  &81.95±5.45  &71.28±1.03  &70.62±2.49  &53.18±2.00   &50.86±0.80  &49.51±1.44     \\
\midrule
PC-GNN    &94.92±0.45  &83.36±0.39  &86.26±0.47  &80.90±0.09  &78.11±0.27  &67.36±0.43  &56.66±0.35   &44.03±3.88  &46.80±1.55    \\
FRAUDER   &93.27±0.62  &88.21±1.36  &89.24±0.35  &72.84±2.78  &62.09±4.56  &59.32±2.18  &51.37±1.22  &52.09±2.09  &48.16±0.93    \\
ConsisGAD  &93.20±0.32  &75.29±0.69  &89.96±0.69  &90.54±0.47  &62.46±1.64  &76.36±0.15  &53.79±0.84   &15.00±4.21  &\underline{51.42±0.20}     \\
\midrule
\textbf{\modelname}    &\textbf{96.62±0.04}  &\textbf{91.83±0.01}  &\textbf{91.77±0.51}  &\underline{92.41±0.19}  &\textbf{81.55±1.05}  &\underline{80.08±0.31}  &\textbf{65.12±0.17}    &\textbf{61.18±0.35}  &\textbf{52.34±1.13}      \\

\bottomrule
\end{tabular}
}
\caption{Performance Comparison on Amazon, YelpChi, and Comp. ``-": out of memory; Bold: the best of baselines; Underline: runner-up.}
\label{tab:summaryResult}

\end{table*}

\begin{figure*}[ht] 
    \subfigure[Amazon.]{    \begin{minipage}{0.32\textwidth}
        \centering      
        \includegraphics[width=1\textwidth, height=3.3cm]{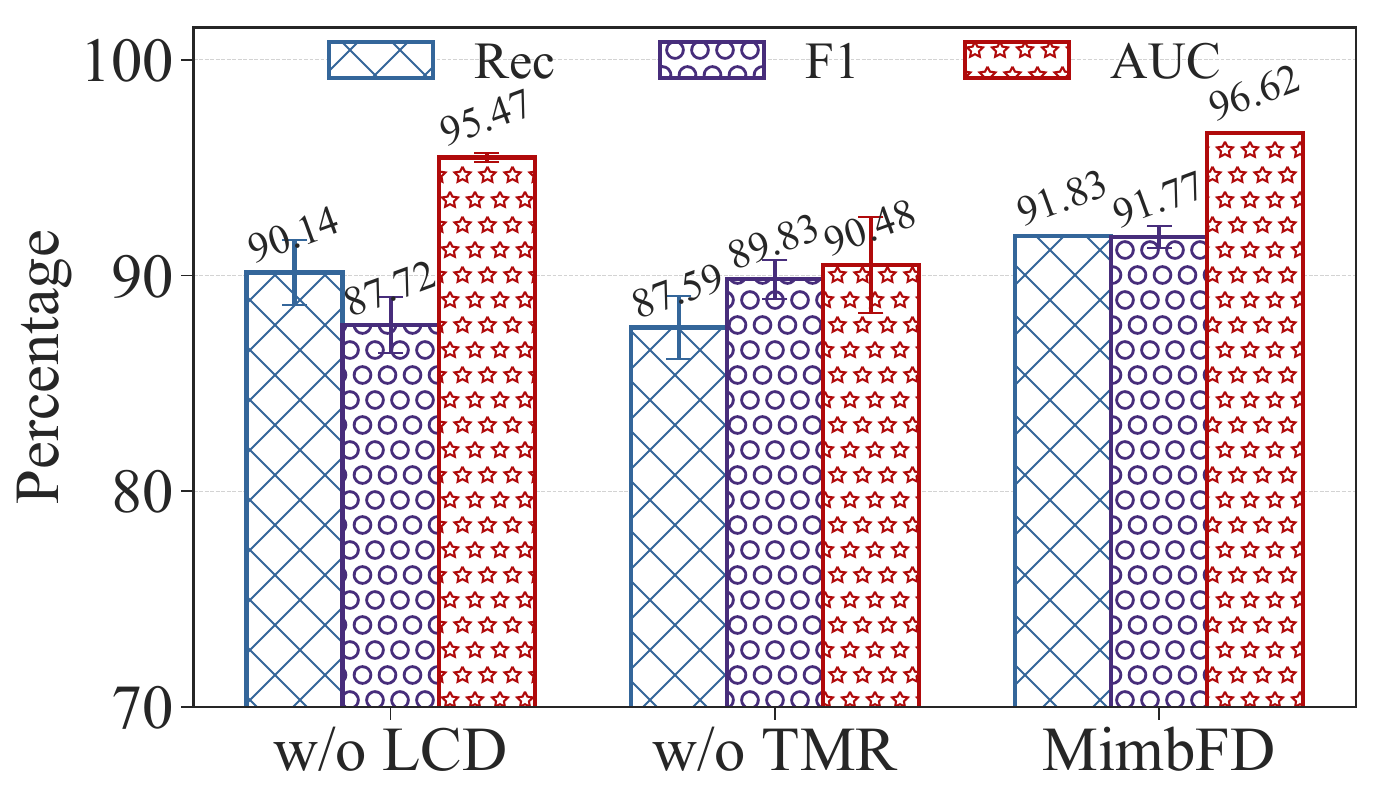}
    \end{minipage}}
    \subfigure[YelpChi.]{  
    \begin{minipage}{0.32\textwidth}
        \centering
        \includegraphics[width=1\textwidth, height=3.3cm]{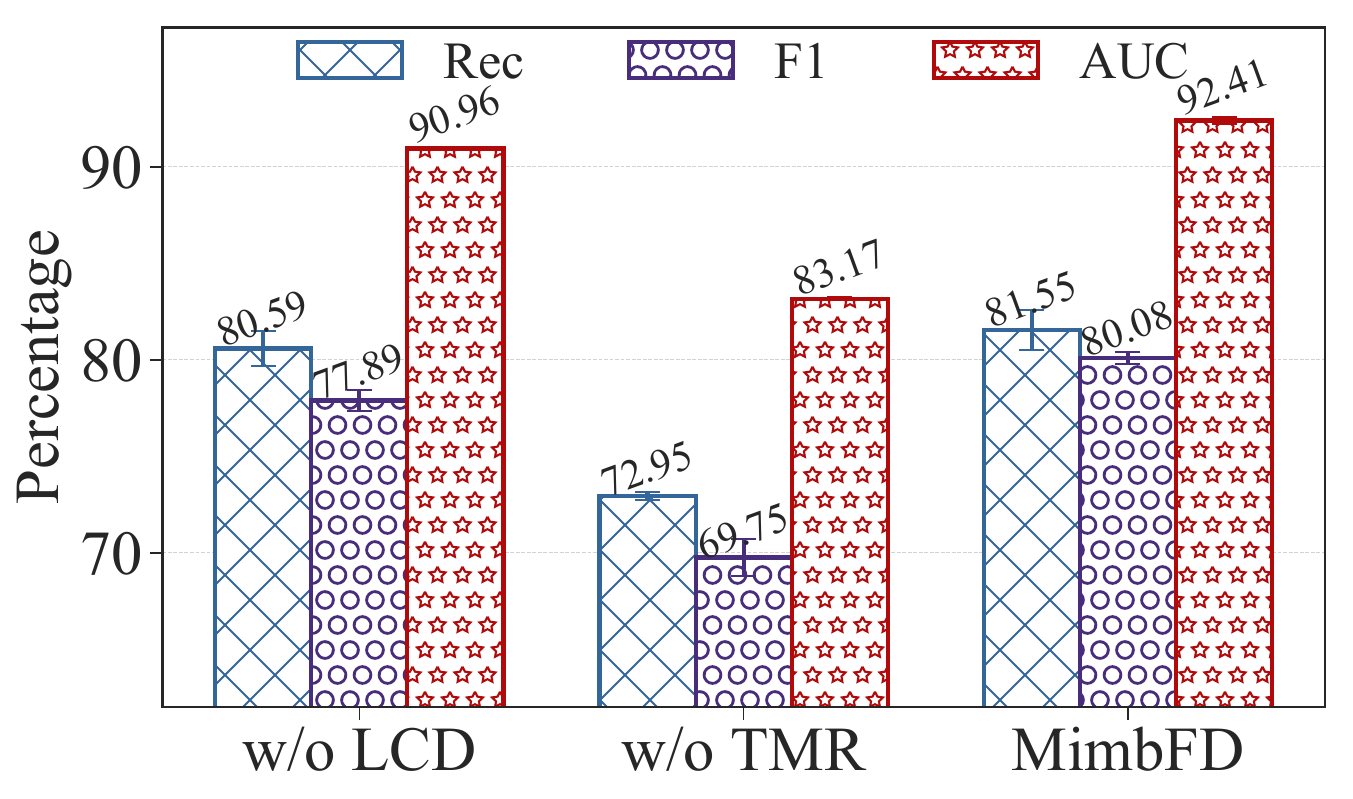}
    \end{minipage}
    }
    \subfigure[Comp.]{  
    \begin{minipage}{0.32\textwidth}
        \centering
        \includegraphics[width=1\textwidth, height=3.3cm]{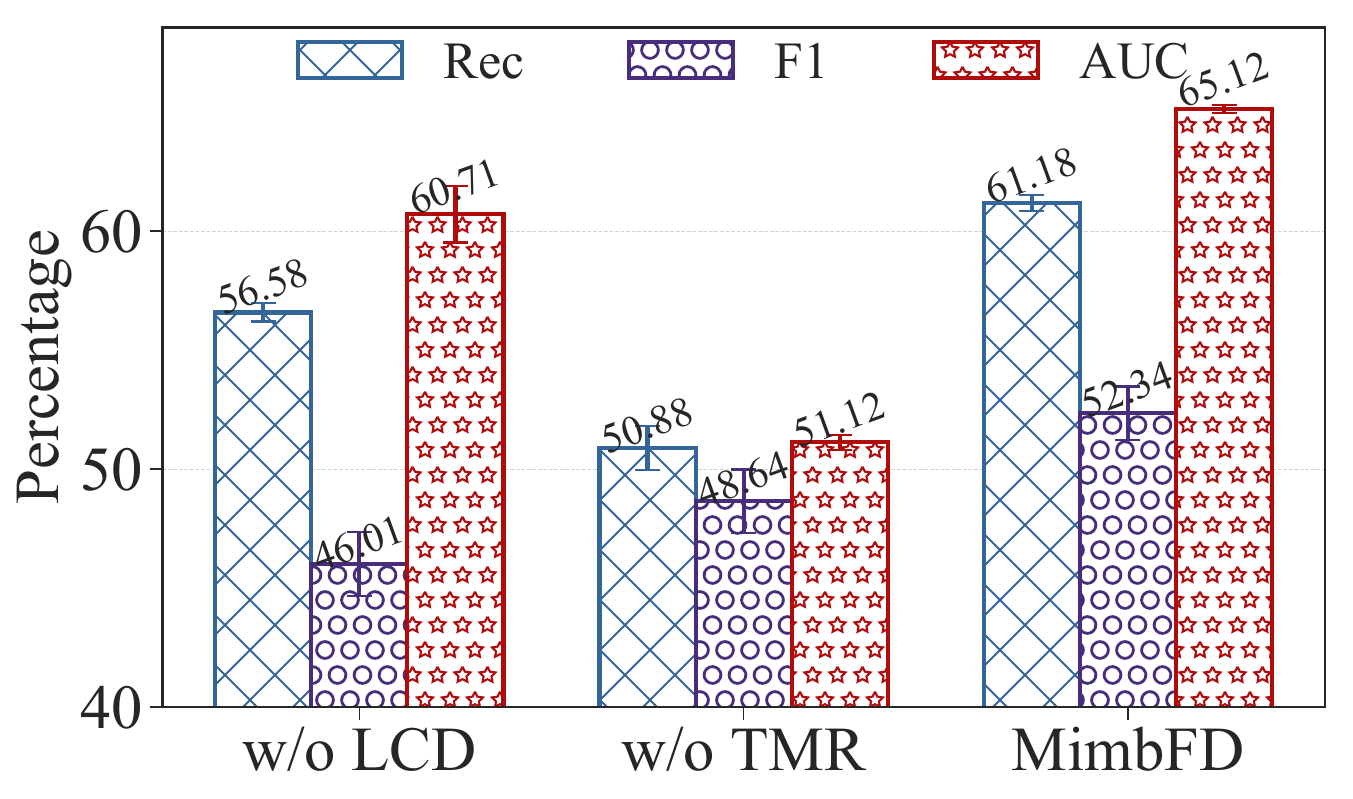}
    \end{minipage}
    }
\centering
\caption{The ablation analysis on Amazon, YelpChi, and Comp.}
\label{fig:Ablation}
\end{figure*}

\section{Experiments}
In this section, we conduct experiments to verify the validity of {\modelname}. 
Specifically, we aim to answer the following research questions:

\textbf{RQ1}: How does our model performance compare to existing state-of-the-art baselines?  
\textbf{RQ2}: How do the modules of TMR and LCD benefit the prediction?
\textbf{RQ3}: How does the {\modelname} perform with different hyperparameters? 
\textbf{RQ4}: Is the {\modelname} able to find fraudsters while still maintaining close ties within the two groups?
\textbf{RQ5}: Can the {\modelname} capture supervisory messages about fraudsters across different levels of imbalance settings?

\subsection{Experimental Setup}

\textbf{Dataset.} 
Three multi-relation graph fraud datasets, YelpChi (Rayana and Akoglu 2015), Amazon (McAuley and Leskovec 2013), and Comp~\cite{Comp} are used.

\noindent \textbf{Baseline.} 
We compare with several state-of-the-art GNN-based methods:
$(1)$ Traditional GNNs: GCN~\cite{GCN}, GAT~\cite{GAT}, and GraphSAGE~\cite{Graphsage}. 
$(2)$ Topological imbalance models: ReNode~\cite{ReNode}, and TAM~\cite{TAM}.
$(3)$ General graph-based FD: CARE-GNN~\cite{CARE-GNN}, GAGA~\cite{GAGA}, and COFD~\cite{COFD}. 
$(4)$ Class imbalance FD: PC-GNN~\cite{PC-GNN}, FRAUDER~\cite{FRAUDRE}, and ConsisGAD~\cite{ConsisGAD}.

\noindent \textbf{Experimental Setting}. 
In our experiments, Adam is chosen as the optimizer. We implement our method through Pytorch and DGL. 
For the dataset split, we divide it into three parts: training set, validation set, and test set, with ratios of 4:2:4, respectively, following common settings. 
For GCN, GAT, GraphSage, and ReNode, we convert multi-relational graphs to isomorphic graphs as input. For fraud detection methods, we use publicly available source code and input multi-relational graphs.

\noindent \textbf{Evaluation Metric}. 
Given the imbalance in the dataset, three widely used metrics, AUC, Recall, and F1-score are used.

\begin{figure*}[ht]
    \subfigure[Amazon.]{    \begin{minipage}{0.32\textwidth}
        \centering      
        \includegraphics[width=1\textwidth, height=3.3cm]{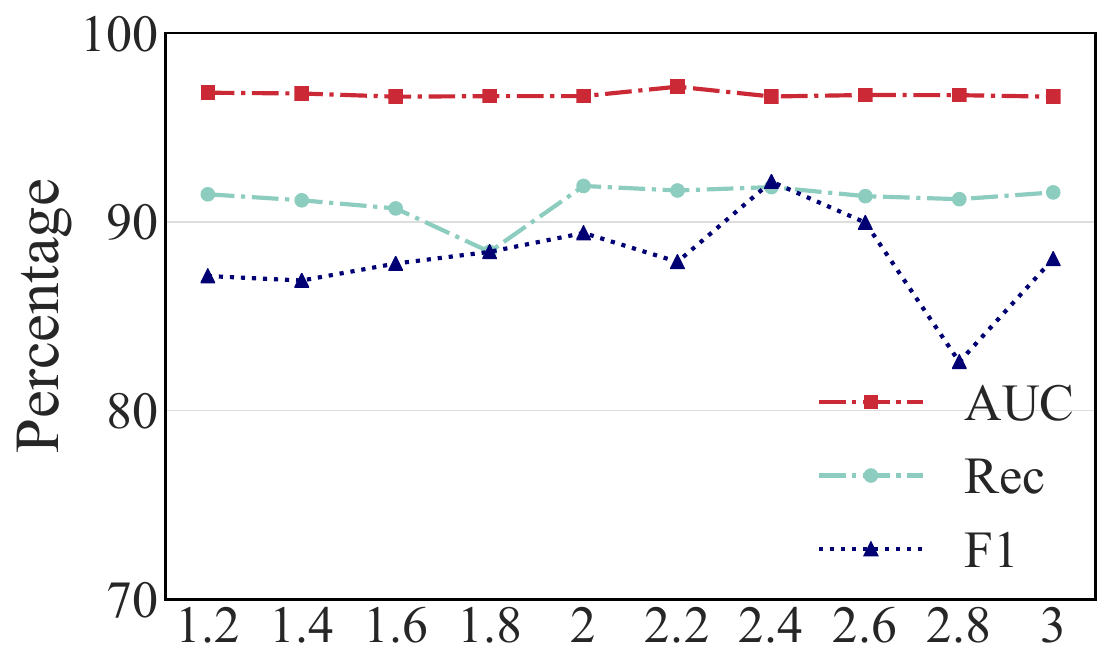}
    \end{minipage}}
    \subfigure[YelpChi.]{  
    \begin{minipage}{0.32\textwidth}
        \centering
        \includegraphics[width=1\textwidth, height=3.3cm]{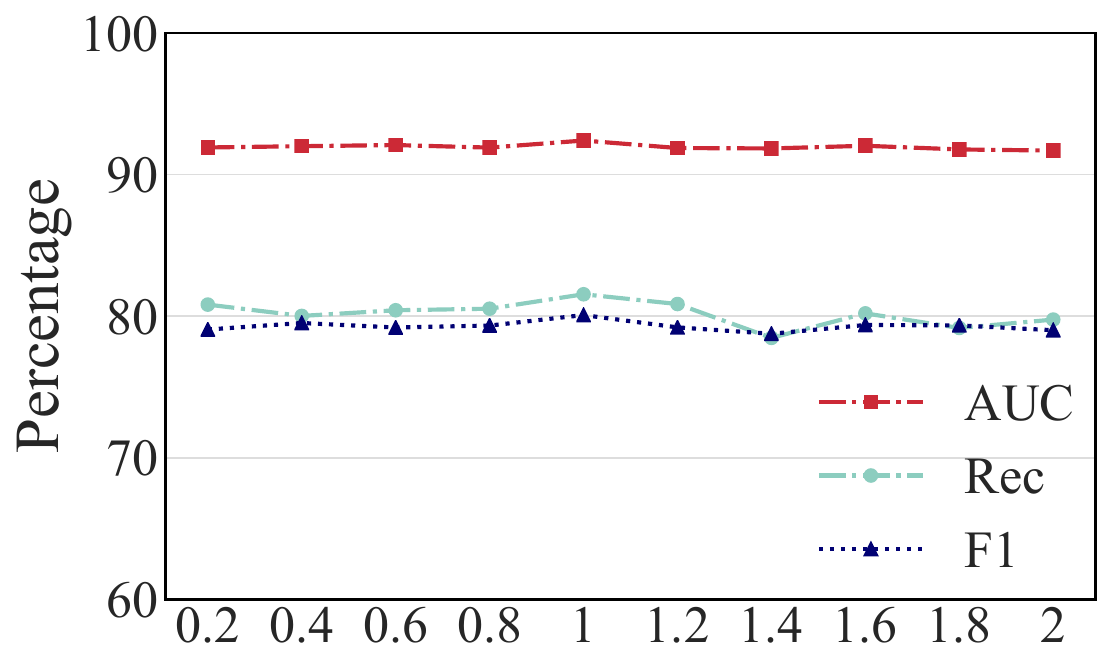}
    \end{minipage}
    }
    \subfigure[Comp.]{  
    \begin{minipage}{0.32\textwidth}
        \centering
        \includegraphics[width=1\textwidth, height=3.3cm]{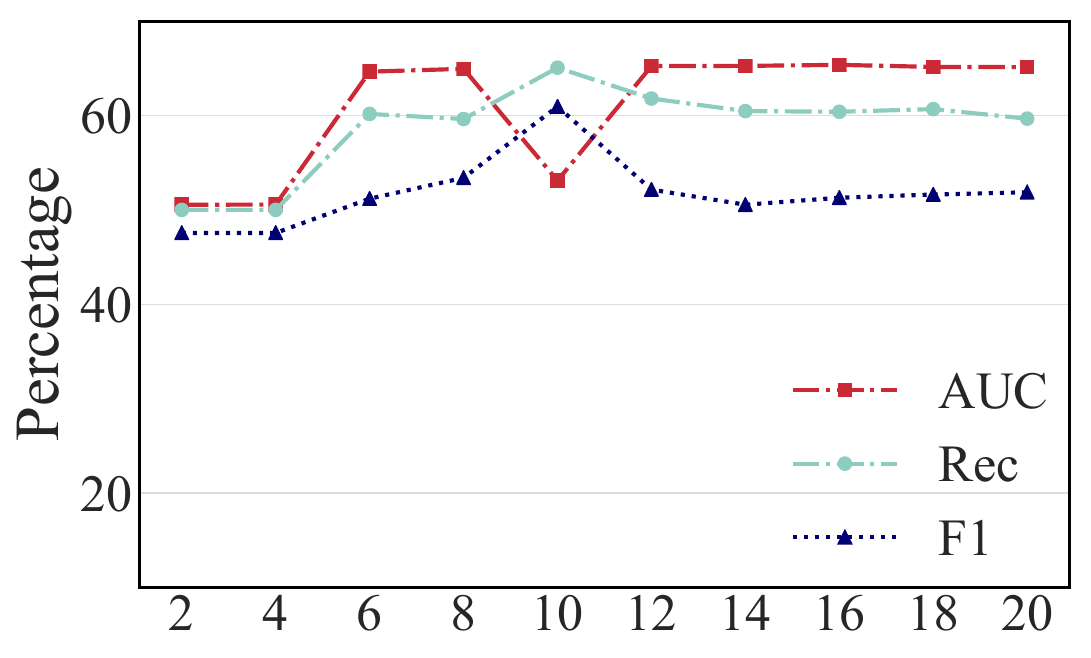}
    \end{minipage}
    }
\centering
\caption{The hyperparameter analysis on Amazon, YelpChi, and Comp.}
\label{hyperparameter}
\end{figure*}

\begin{figure}[ht]
    \centering
    \subfigure[GCN.]{
        \begin{minipage}{0.22\textwidth}
            \centering
            \includegraphics[height=2.5cm]{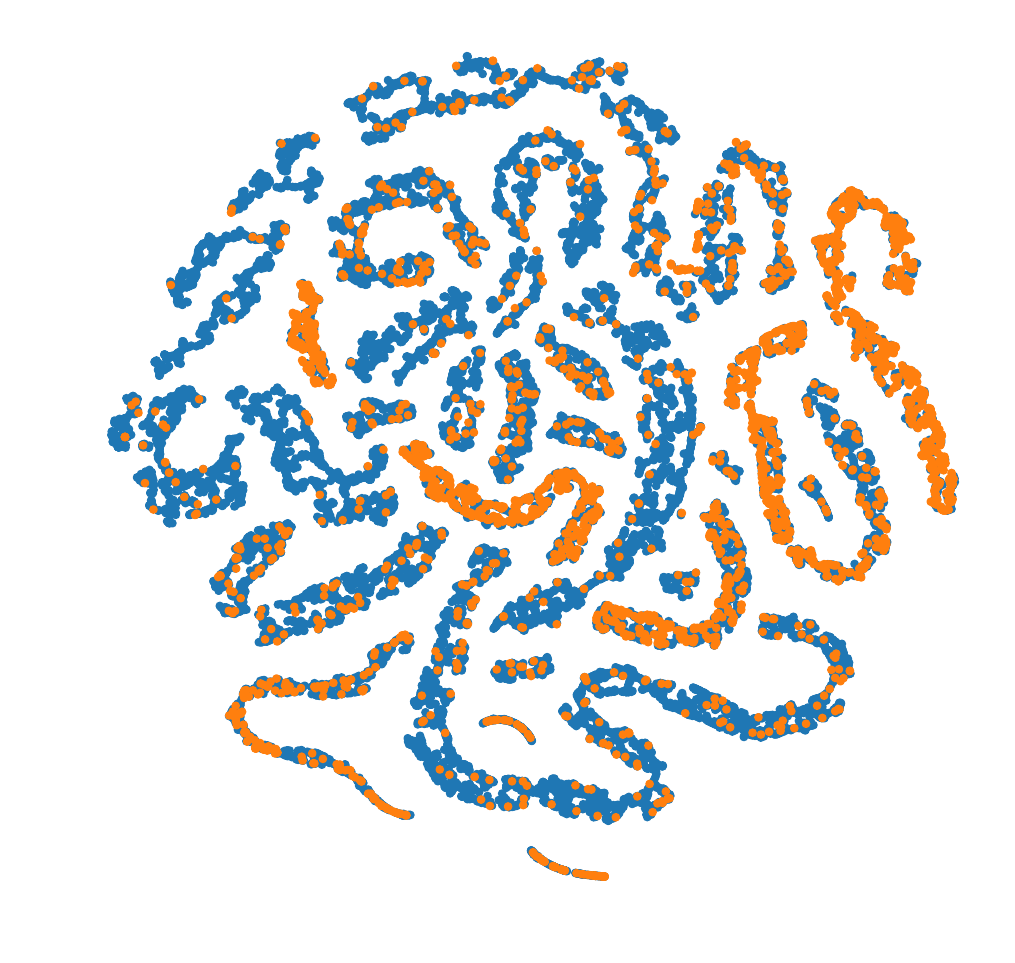}
        \end{minipage}
    }
    \subfigure[GAGA.]{
        \begin{minipage}{0.22\textwidth}
            \centering
            \includegraphics[height=2.5cm]{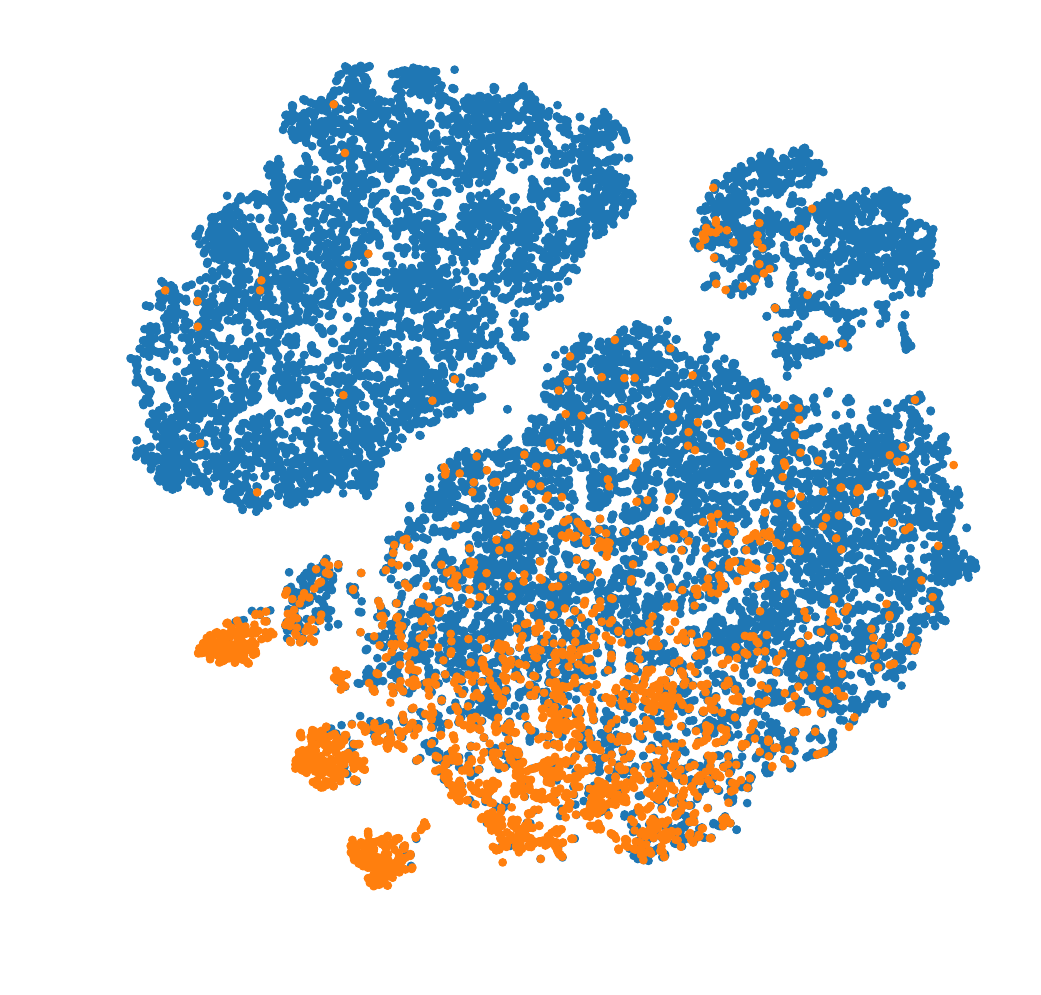}
        \end{minipage}
    }
    \subfigure[ConsisGAD.]{
        \begin{minipage}{0.22\textwidth}
            \centering
            \includegraphics[height=2.5cm]{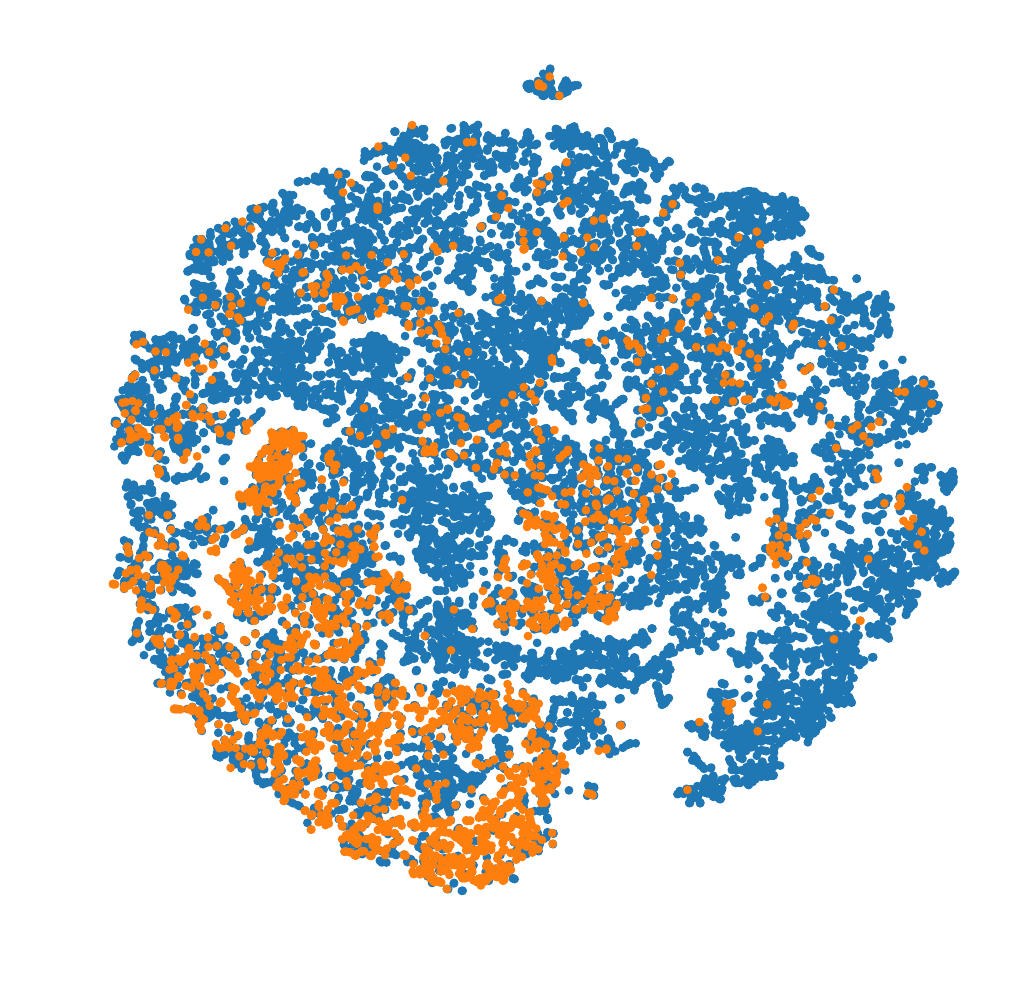}
        \end{minipage}
    }
    \subfigure[{\modelname}.]{
        \begin{minipage}{0.22\textwidth}
            \centering
            \includegraphics[height=2.5cm]{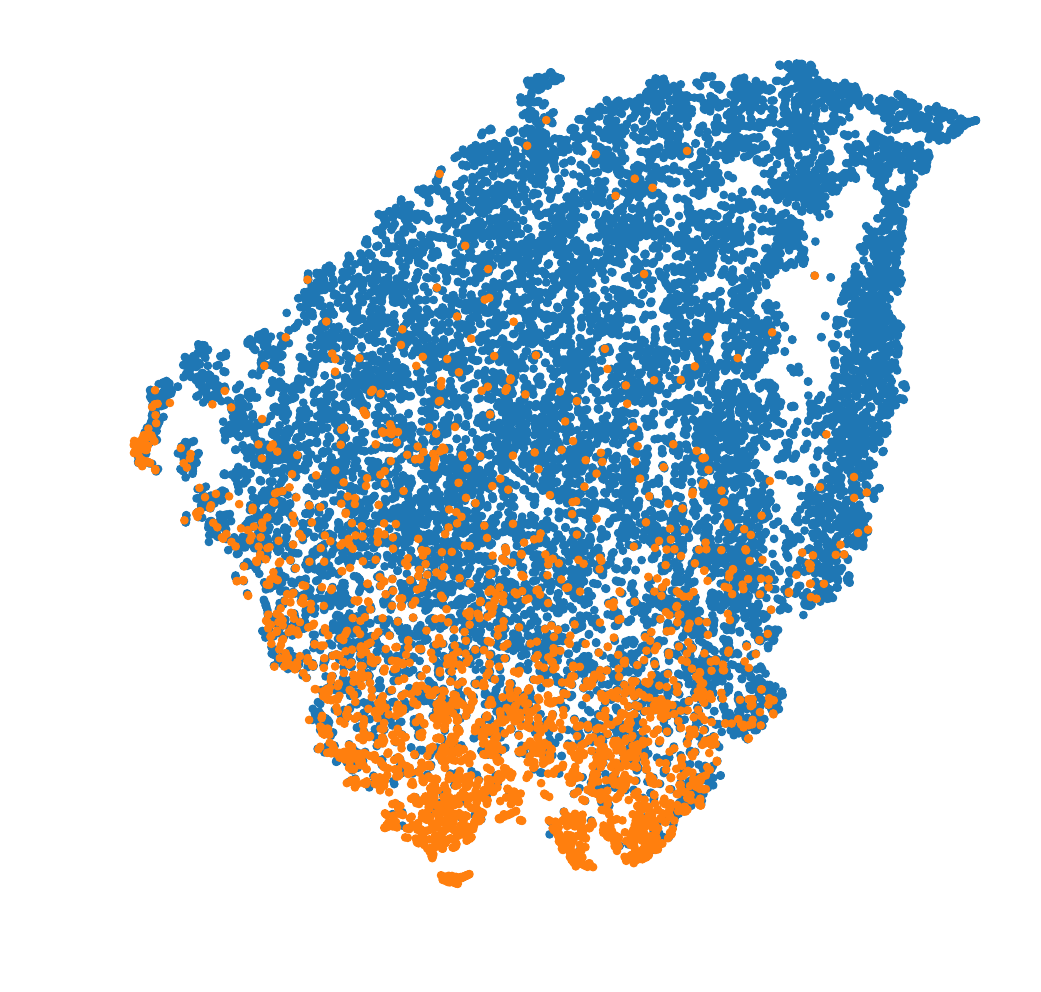}
        \end{minipage}
    }
    \caption{Visualization analysis on YelpChi.}
    \label{Visualization}
\end{figure}
\subsection{Performance (RQ1)}\label{}
To answer RQ1, we compare our method with the state-of-the-art approaches. The results are shown in Tab.~\ref{tab:summaryResult}. 
Due to the homophily assumption, the topological methods poorly handle fraud supervision signals, hindering balanced information propagation. The relatively lower performance of FRAUDER and ConsisGAD, which focus on class imbalance, as well as CARE-GNN, a general graph-based fraud detector, can be attributed to their emphasis on local tuning. This highlights the fact that learning node representations using only first-order neighborhood information is insufficient to support effective model judgments. 
In contrast, fraud detectors like PC-GNN, which address class imbalance, and general graph-based fraud detectors like GAGA and COFD, are more competitive, suggesting that the consideration of long-distance information is crucial.

Our method almost outperforms the baseline performance in terms of overall performance.
Its strong performance on Comp, which contains more conflicting supervisory signals, highlights its effectiveness in balancing supervision coverage and amplifying fraud-related information.

\subsection{Ablation Analysis (RQ2)}\label{Ablation}
To answer RQ2, we conduct a series of ablation studies on certain modules in {\modelname} on three datasets, as shown in Fig.~\ref{fig:Ablation}. The variants tested are: w/o LCD and w/o TMR.
The w/o TMR module focuses on distinguishing between the information in node representations and can capture attempts by fraudsters to obfuscate messages that reveal their identity. This helps alleviate the class imbalance problem in fraud detection to some extent.
On the other hand, the w/o LCD module, which captures global supervisory information across the complex network structure, enables adaptive learning of supervisory messages by the nodes. It has been shown to mitigate the issue of uneven signal propagation caused by the fraudster's long-distance camouflage.

{\modelname} outperforms both variants across all metrics. 
The performance of both variants on Comp confirms that long-distance supervision information provides critical signals for effective identification.

\subsection{Hyperparameter Analysis (RQ3)}
To answer RQ3, we evaluate the impact of the hyperparameter $\eta$, which balances the interaction between the TMR and LCD modules, on the performance of {\modelname} across different datasets, as shown in Fig.~\ref{hyperparameter}. 
The results show that all metrics are relatively stable on the Amazon and YelpChi datasets, whereas the metrics fluctuate more on the Comp dataset. This is due to the smaller data volume of Comp, where the supervisory information received by the nodes leads to significant conflicts, thus requiring more training for the LCD module.

Overall, the trends are roughly consistent across the datasets. When  $\eta$ is small, the model struggles to capture key signals in node representations, resulting in lower performance. As $\eta$ increases, the model trains more effectively, leading to improved judgment. However, when $\eta$ becomes too large, excessive bias correction occurs, as the model places too much emphasis on addressing weak signals while neglecting the connection between the true key information and the node identity.

\subsection{Visualization (RQ4)}
To answer RQ4, we visualize the node embeddings of different methods on the YelpChi dataset. The results are shown in Fig.~\ref{Visualization} and the appendix. 
GCN only considers the homophily problem in node classification, which is insufficient for fraud detection, leading to poor separation results. 
GAGA is more competitive, achieving significant inter-class separation. This confirms that leveraging distant neighbors to exploit high homophily to resist camouflage is effective. ConsisGAD addresses the class imbalance problem in fraud detection, but does not fully utilize information from various types of nodes.

In contrast, our method generates more distinct separation effects and better intra-class cohesion. This demonstrates that {\modelname} effectively adjusts the supervisory information within the global topology and amplifies the capture of key signals from fraudsters, allowing it to uncover the fraudster's camouflaged behavior and identity more efficiently.

\subsection{Case Study (RQ5)}
\begin{table}
\resizebox{\linewidth}{!}{
\centering
\begin{tabular}{c|cccc}
    \toprule
       $\rho$  &GCN & GAGA & ConsisGAD & {\modelname}   \\
    \midrule

    5    &79.12±3.42   & 84.39±6.86        & 75.52±1 15     & \textbf{95.90±0.38}     \\

    \midrule            

    10     &82.17+0.65   & 87.95±0.70     & 76.57±1.11      & \textbf{91.83±0.01}    \\

    \midrule

    20       &81.01+0.54    & 84.62±1.48       & 76.16±0.55    & \textbf{92.13±0.66}       \\

    \bottomrule
\end{tabular}
}
\caption{The case study on Amazon. Bold: the best of the baselines.}
\label{tab:case}
\end{table}

To answer RQ5, we use Recall as the evaluation metric, as shown in Tab.~\ref{tab:case}. We set up three imbalance scenarios based on the original Amazon dataset, where about 10\% are fraudsters. We adjusted the ratio of fraudsters to benign users, $\rho$, to $1:5$ and $1:20$ ($\rho=5$ and $\rho=20$) to represent moderately to extremely imbalanced data.

The results show that {\modelname} performs well across all imbalance settings. At $\rho = 5$, where the two classes are easily confused, {\modelname} achieves the best performance, highlighting its strength in mitigating the bottleneck of limited supervision transmission.
Other baselines perform poorly at $\rho=20$ because they cannot enhance their representations by obtaining long-distance fraud information.
\section{Conclusions}

We propose {\modelname} to tackle fraud information imbalance from a dual perspective.
For topological imbalance, we balance supervisory message reception to maximize information utilization.
For class imbalance, we enhance identity gain by extracting key signals from node representations.
Extensive experiments demonstrate that our model effectively mitigates information imbalance.

\clearpage
\bibliographystyle{named}
\section*{Acknowledgments}
The corresponding authors are Li-e Wang and Xingcheng Fu. This work is supported in part by the National Natural Science Foundation of China (Nos. 62262003, 62462007 and U21A20474), the Guangxi Science and technology project (Guike AA22068070), the Guangxi Bagui Youth Talent Training Program, Guangxi Collaborative Innovation Center of Multisource Information Integration and Intelligent Processing and the Key Lab of Education Blockchain and Intelligent Technology, Ministry of Education (EBME24-01).

\section*{Contribution Statement}
% † Co-first Authors.
Co-first authors are Yudan Song and Yuecen Wei, marked with †.

\bibliography{references}

\appendix
\clearpage

\section{Experiments}

\subsection{Dataset} \label{sec:Dataset}
The statistics of the dataset are shown in table~\ref{tab:dataset}. In addition, a description of the dataset is as follows:
\begin{itemize}
    \item \textbf{YelpChi:} collects reviews (spam) and recommendations (legitimate) on Yelp platforms about hotels and restaurants, with the following three main relationships: 1) R-U-R: reviews posted by the same user are connected, 2) R-T-R: reviews posted under the same month based on the same product are connected, and 3) R-S-R: reviews under the same product with the same star rating are connected.
    \item \textbf{Amazon:} records reviews of products used under the musical instruments category. There are three main relationships: 1) U-P-U: user reviews under at least one of the same product are connected, 2) U-S-U: users with at least one review of the same star rating in a week are connected, and 3) U-V-U: users who rank in the top 5$\%$ of all users in terms of similarity of the text of their mutual reviews are connected (as measured by TF-IDF).
    \item \textbf{Comp:} records the financial statements of Chinese companies. It contains three relationships: 1) C-I-C: connects companies with investment relationships, 2) C-P-C: connects companies and their disclosed customers, and 3) C-S-C: connects companies and their disclosed suppliers.
\end{itemize}

\begin{table}[ht]
    \centering
    \begin{tabular}{lrrrrr}
        \toprule
        Dataset   &\#Nodes & Fraud(\%) & Relation & \#Edges   \\
        \midrule
                   &         &        & R-U-R    &98,630      \\
        YelpChi    &45,954   & 14.5\%        & R-T-R     &1,147,232       \\
                    &        &        & R-S-R     &6,805,486          \\
        \midrule            
                    &        &        & U-P-U    &351,216        \\
        Amazon     &11,944   &9.5\%        & U-S-U      &7,132,958          \\
                   &         &        & U-S-U     &2,073,474            \\
        \midrule
                   &         &        & C-I-C    & 5,686       \\
        Comp       &5,317    &10.5\%        & C-S-C     &760        \\
                    &        &        & C-P-C     & 1,043      \\
        \bottomrule
    \end{tabular}
    \vspace{-0.5em}
    \caption{The Statistics of Datasets. \# represents quantity.}
    \label{tab:dataset}
\end{table}

\subsection{Baseline}

A description of the backbone model used is as follows:

\begin{itemize}
    \item \textbf{GCN}~\cite{GCN}: a graph convolutional network that generates final node embeddings by aggregating information from first-order neighbors. 
\end{itemize} 
\begin{itemize}
   \item\textbf{GAT}~\cite{GAT}: a graph attention network with neighbor weights computed by the attention mechanism and eventually aggregated based on the weights.
\end{itemize} 
\begin{itemize}
    \item\textbf{GraphSAGE}~\cite{Graphsage}: an inductive framework that samples and aggregates neighboring nodes to generate node representations.
\end{itemize} 
\begin{itemize}
    \item\textbf{CARE-GNN}~\cite{CARE-GNN}: a camouflage-resistant graph neural network for neighbor selection using label similarity metric and reinforcement learning.
\end{itemize}
\begin{itemize}
    \item\textbf{PC-GNN}~\cite{PC-GNN}: a fraud detection method that balances node distribution using labels and neighborhood samplers.
\end{itemize}
\begin{itemize}
    \item\textbf{FRAUDER}~\cite{FRAUDRE}: a fraud detection effort that aggregates the differences between neighbors and handles class imbalances.
\end{itemize}
\begin{itemize}
    \item\textbf{GAGA}~\cite{GAGA}: a fraud detection method based on a Transformer capturing semantic information to deal with the low congruence problem.
\end{itemize}
 \begin{itemize}
     \item\textbf{COFD}~\cite{COFD}: a fraud detector to detect collaborative fraud by focusing on the comparison of first and second order node pair structures.
 \end{itemize}
 \begin{itemize}
     \item\textbf{ConsisGAD}~\cite{ConsisGAD}: a fraud detector that mitigates class imbalance by performing domain consistency augmentation for homophilic distribution of normal and fraudulent nodes.
 \end{itemize}

 \subsection{Matrics}
The meanings of the indicators we use are as follows: 
 \begin{itemize}
  \item \textbf{Recall} indicates the proportion of correctly predicted positive instances among all actual positive instances in the sample; a higher value implies better performance. 
   \item \textbf{AUC} provides a reliable evaluation of the classifier's overall performance, particularly effective in patterns with imbalanced samples.
    \item \textbf{F1-score} represents the harmonic mean of precision and recall, with a higher value indicating increased precision and recall while minimizing variance.
\end{itemize}

\subsection{Additional results}

T-Finance\footnote{Jianheng Tang, et al. Rethinking Graph Neural Networks for Anomaly Detection. ICML'22.} is a single-relationship fraud graph designed to identify anomalous accounts within a trading network.
To assess the method’s utility on a larger dataset, we compared it against selected baselines.
As shown in ~\ref{tab:AddsummaryResult}, {\modelname} consistently outperforms the baselines, demonstrating its ability on the larger dataset effectively mitigate the imbalanced propagation of supervisory messages in fraud detection scenarios.

\begin{table}[ht]
    \centering
    \begin{tabular}{cccc}
        \toprule
        Model    &AUC & Recall &F1    \\
        \midrule  
        GCN    &88.33±1.06   & 51.24±9.87       &61.37±3.71           \\
  
        \midrule            

        ReNode     &42.79±28.85   &79.40±5.84        & 46.5±8.2              \\

        \midrule

        CARE-GNN        &91.76±0.32    &78.11±0.36       & 73.54±0.65           \\

        GAGA        &95.53±0.45    &87.78±0.58        & 90.65±1.25           \\
        \midrule

        PC-GNN        &-    &-        & -           \\

        ConsisGAD         &96.61±0.07    &78.32±1.59        & 91.51±0.05          \\  
        \midrule        
        \textbf{\modelname}        &\textbf{96.69±0.44}    &\textbf{88.67±0.39}        & \textbf{92.05±0.33}         \\

        \bottomrule
    \end{tabular}
    \vspace{-0.5em}
    \caption{Performance Comparison on T-Finance. Bold: the best of baselines.}
    \label{tab:AddsummaryResult}
\end{table}

\end{document}